\documentclass[11pt]{article}

\usepackage[final]{latex/acl}

\usepackage{times}
\usepackage{latexsym}
\usepackage{booktabs}
\usepackage{multirow}
\usepackage{algorithm}
\usepackage{amsmath}
\usepackage{amssymb}
\usepackage{algpseudocode}

\usepackage[T1]{fontenc}

\usepackage[utf8]{inputenc}

\usepackage{microtype}

\usepackage{inconsolata}

\usepackage{graphicx}

\title{From Rollouts to Recipes: Self-Contained Post-Training for LLMs}

\author{
  \textbf{Yifei Li}$^{1,3,4}$ \quad
  \textbf{Lingling Zhang}$^{1,2,3}$\thanks{Corresponding author.}\quad
  \textbf{Muye Huang}$^{1,3,4}$ \quad
  \textbf{Zihan Ma}$^{1,3}$ \\
  \textbf{Jiashuai Liu}$^{1,3}$ \quad
  \textbf{Jun Liu}$^{1,2,3}$ \\[2mm]
  $^{1}$School of Computer Science and Technology, Xi'an Jiaotong University, China \\
  $^{2}$National Engineering Research Center for Visual Information and Applications \\
  $^{3}$Shaanxi Province Key Laboratory of Big Data Knowledge Engineering \\
  $^{4}$Zhongguancun Academy, Beijing, China\\
  \texttt{yifeilee@stu.xjtu.edu.cn, zhanglling@xjtu.edu.cn}
}

\begin{document}
\maketitle
\begin{abstract}
Post-training large language models usually applies a single training recipe to all samples, even though the model's own rollouts reveal different sample-level learning states. We propose Self-Routing, a behavior-conditioned post-training framework that uses rollout correctness and confidence to decide how each sample should be optimized. Depending on its behavior state, a sample is routed to GRPO, on-policy self-distillation, regularization, or skipping, allowing training to adapt without external teachers, extra annotations, or additional sampling. Experiments on mathematical reasoning across Qwen3 and Qwen3.5 backbones show that Self-Routing consistently improves over uniform GRPO, uniform OPSD, fixed mixtures, and simpler routing baselines. Further analyses show that the routing distribution changes over training and reduces unnecessary updates on low-signal or already stable samples. 
\end{abstract}

\section{Introduction}
\label{sec:introduction}

Post-training is central to improving the reasoning ability of large language models (LLMs) on verifiable tasks such as mathematical reasoning, code generation, and program repair. In these settings, verifier-based reinforcement learning can use outcome rewards, such as answer correctness, test pass rates, or patch validity, without relying on large-scale chain-of-thought (CoT) annotations or human process supervision~\citep{uesato2022solving,le2022coderl,liu2023rltf,gehring2024rlef,wei2025swerl}. This makes naive RL / RLVR attractive when CoT data are scarce, noisy, or mismatched with the target model~\citep{guo2025deepseekr1,yu2026dapo}. Prior work has improved verifier-based post-training through reward design, advantage estimation, regularization, sampling, distillation, and hybrid objectives~\citep{schulman2017ppo,shao2024deepseekmath,guo2025deepseekr1,yu2026dapo}. Yet most methods still follow a global training recipe: they apply the same optimization mechanism, or a fixed objective mixture, to all samples in the dataset.

\begin{figure}[t]
  \includegraphics[width=\columnwidth]{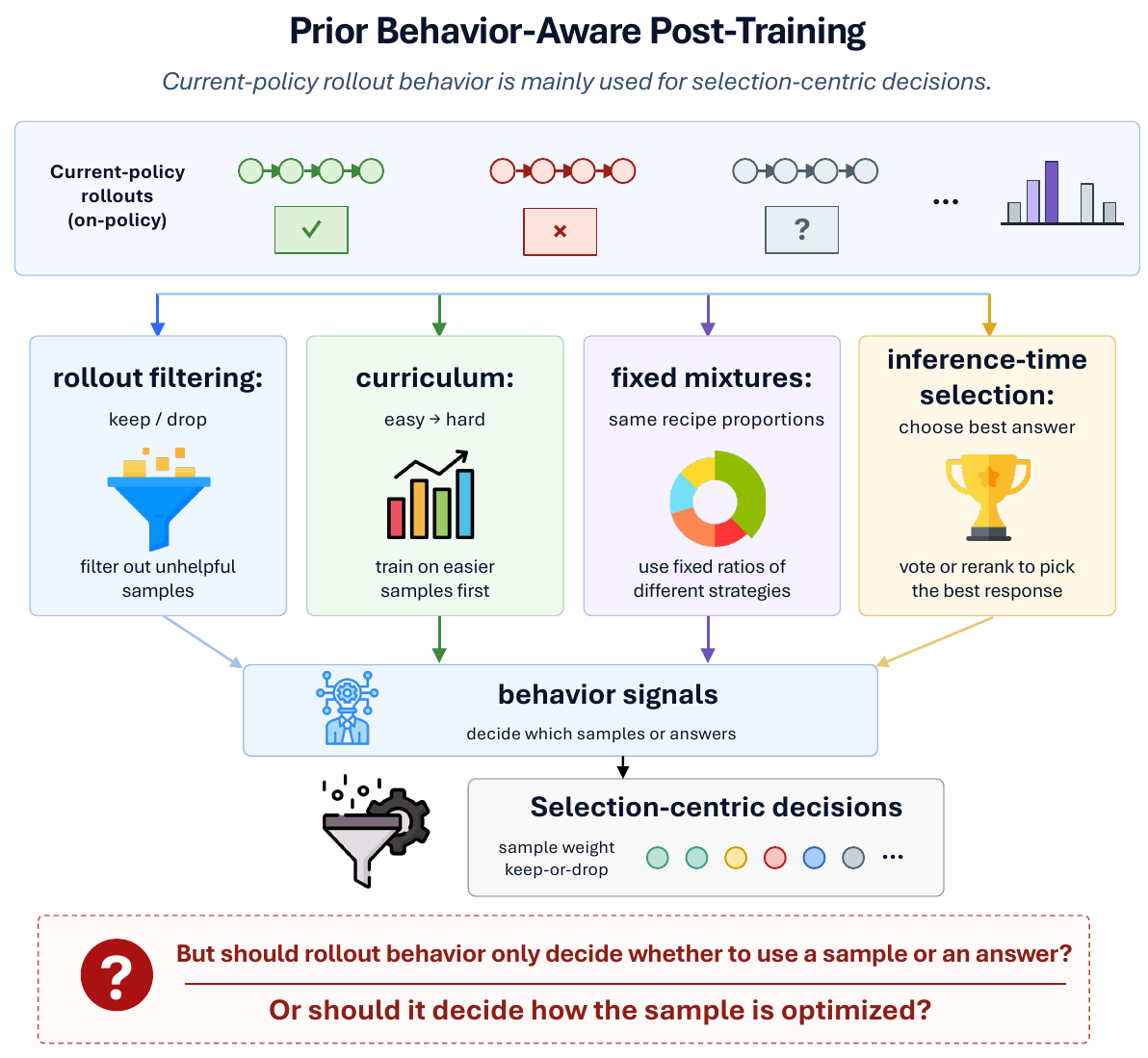}
  \caption{Prior behavior-aware post-training mainly uses rollout signals for selection, such as filtering samples, scheduling curricula, mixing fixed recipes, or choosing inference-time answers.}
  \label{fig:overview}
\end{figure}

This global view overlooks the fact that rollouts already contain sample-level behavioral signals. Existing methods partially exploit this information: all-correct / all-wrong filtering skips samples whose rollouts provide no relative advantage signal~\citep{yu2026dapo,shao2024deepseekmath}, while entropy-, confidence-, or consistency-based voting and reranking can improve final answer quality without updating model parameters~\citep{wang2022selfconsistency,weng2022selfverification}. These observations suggest that rollout behavior reflects the current learning state of each sample. Importantly, such signals are available during rollout generation itself, without extra annotation, external evaluators, or redundant inference.

A sample's training value is therefore not a static property such as quality or difficulty. It depends on the model's current rollout behavior and on the optimization signal applied to that sample. A sample with uniformly failed rollouts may provide little gradient for verifier-based RL, while on-policy distillation can still offer dense guidance on states visited by the current model. A sample with mixed correct and incorrect rollouts may be informative for RL, while a reliably solved sample may be better treated with a conservative objective. Samples that repeatedly fail with high confidence may be better skipped temporarily. We therefore view post-training as a process from rollouts to recipes: rollouts expose local learning states, and these states can be converted into sample-level training recipes that decide both whether and how each sample should be optimized.

Based on this view, we propose self-contained behavior-conditioned routing. Unlike global recipes that fix an optimization mechanism and apply it uniformly, our approach assigns training actions according to each sample's rollout behavior under the current model. Unlike distillation from external teachers, offline CoT traces, or fixed reference trajectories, our signals come from the model's own on-policy rollouts, reducing the risk of mode-distribution mismatch between the imitated trajectories and the states actually visited by the current policy~\citep{agarwal2023onpolicy,zhao2026self}. This forms a self-contained loop: the model reveals its learning states through rollouts, and training uses those states to decide subsequent updates.

In summary, our contributions are as follows:
\begin{itemize}
    \item We identify a structural limitation of verifier-based post-training: global recipes cannot adapt to heterogeneous sample-level learning states revealed by current-policy rollouts.
    \item We propose self-contained behavior-conditioned routing, which turns rollout signals naturally produced during training into sample-level recipes without extra annotation, external evaluators, or additional sampling.
    \item We empirically show that different rollout behavior states benefit from different optimization mechanisms, and that behavior-conditioned routing improves over uniform GRPO, uniform OPSD, fixed objective mixtures, and filtering- or curriculum-based baselines.
\end{itemize}

\section{Related Work}
\label{sec:related_work}

\paragraph{Verifier-Based Post-Training.}
Verifier-based post-training has been widely studied for verifiable tasks such as mathematical reasoning, code generation, and program repair, where final answers, unit tests, or execution feedback can provide outcome rewards~\citep{kimi2025k1,hu2025openreasoner,zhao2025absolutezero}. 
Existing work mainly improves the reinforcement learning procedure itself, including PPO/GRPO/DAPO-style objectives, reward shaping, advantage estimation, KL regularization, length control, sampling strategies, and training stabilization~\citep{hu2025openreasoner,kimi2025k1,wang2025entropytokens,xu2025pods}. 
Other methods combine verifier rewards with distillation or supervised signals to mitigate the sparsity of outcome rewards~\citep{zelikman2022star,zhang2025bread,hubotter2026selfdistillation}. 
Relatedly, self-specialized teacher distillation addresses the loss of general capabilities during target-only post-training without requiring a representative replay corpus
\citep{li2026selfspecializedteachersdomainposttraining}.
Beyond mathematical reasoning, related studies have also examined learning and evaluation in more complex agent settings, including long-horizon memory, cross-platform action transfer, and risk under task complexity~\citep{li2026locomo,yan2026maga,ma2025brittle}.
Our work shares the same broad goal of improving model behavior through interaction and feedback, but focuses specifically on verifier-based post-training: rather than designing another global RL objective, we study how rollout states can route samples among different training actions.

\paragraph{Data Selection and Curriculum Learning.}
Data selection and curriculum learning adjust the training distribution according to sample quality, difficulty, loss, uncertainty, reward, or training stage~\citep{wang2025oneshotrlvr,jiang2025vcrl,zhao2025absolutezero}. 
In RLVR, all-correct / all-wrong filtering is a common example: when all rollouts in a group are correct or all are incorrect, the sample provides little relative advantage signal and is skipped or down-weighted~\citep{xu2025pods,jiang2025vcrl}. 
These methods mainly address which samples should be trained, or in what order and frequency they should appear. 
In contrast, our work asks a complementary question: given a sampled prompt and its rollouts, what optimization signal should be applied to it?

\paragraph{Rollout Uncertainty and On-Policy Signals.}
Rollout distributions and model-internal uncertainty signals have been used extensively to adapt inference-time reasoning. 
Self-consistency, majority voting, confidence-based selection, and entropy-aware reranking use such signals for answer selection~\citep{chen2023universal,yao2023tree,jiang2023llmblender,zuo2025ttrl}, while recent work further uses uncertainty and consistency to dynamically allocate reasoning computation~\citep{yan2026mur,xu2026dual}.
On-policy learning and on-policy distillation reduce distribution mismatch by supervising states visited by the current policy rather than relying only on offline trajectories or fixed references~\citep{tan2023gkd}. 
On-policy self-distillation goes one step further: its teacher signal comes from the model itself or its delayed/historical versions, which better matches the model's reasoning style, length bias, and generation distribution than an external teacher~\citep{hubotter2026selfdistillation,zhang2025corewarding}. 
This is especially relevant for reasoning models, where different teachers may solve the same problem through substantially different intermediate patterns. 
We therefore use OPSD as a self-contained primitive for providing dense guidance under sparse rewards, and extend the use of rollout behavior from inference-time adaptation to training-time recipe routing.

\begin{figure*}[t]
    \centering
    \includegraphics[width=\textwidth]{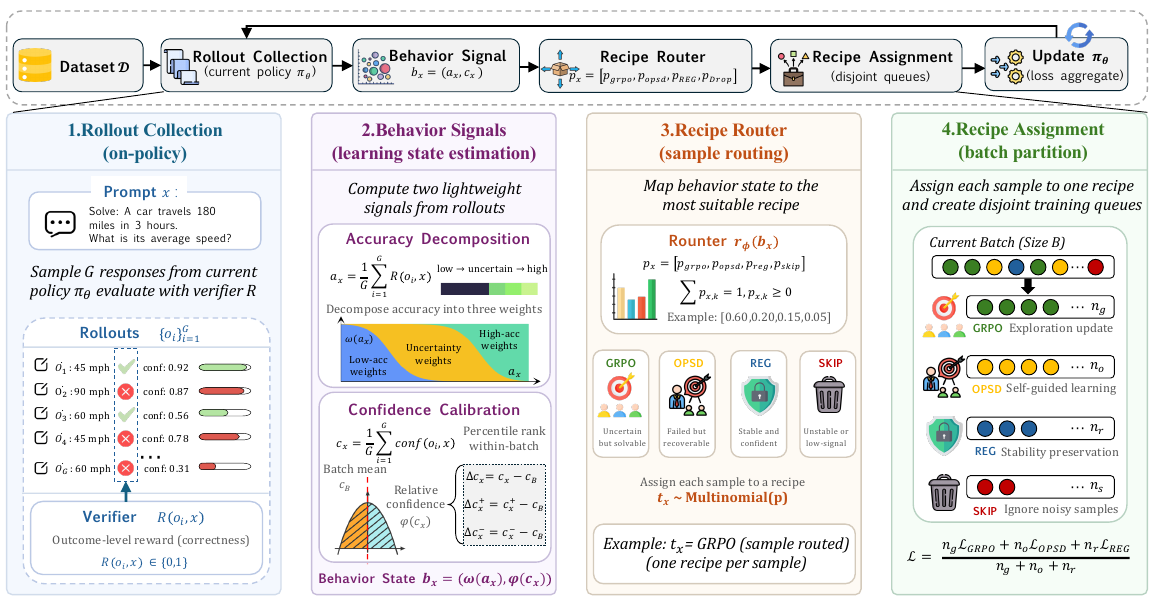}
    \caption{
    Overview of our behavior-aware recipe routing framework. 
    For each training sample, the current policy collects on-policy rollouts, estimates behavior signals from verifier correctness and model confidence, routes the sample to one training recipe, and updates the policy using the aggregated losses from disjoint recipe queues.
    }

    \label{fig:method_overview}
\end{figure*}
\section{Method}

\subsection{Overview}

We study verifier-guided post-training on a fixed training set
\[
\mathcal{D}=\{x_j\}_{j=1}^{N},
\]
where each \(x\) denotes a prompt, problem, or instruction input. Starting from an initial policy \(\pi_{\theta_0}\), the goal is to update the policy on \(\mathcal{D}\) and evaluate it on held-out or related task distributions. GRPO, OPSD, and our method share the same data, rollout budget, and verifier feedback; they differ in how they use these signals to construct training updates.

Figure~\ref{fig:method_overview} illustrates our framework. At each training iteration, the current policy first generates on-policy rollouts for each sample. We then estimate two behavior signals: rollout correctness and model confidence. These signals define a behavior state, which is passed to a recipe router. The router assigns each sample to one of four disjoint queues: GRPO, OPSD, REG, or SKIP. Finally, the active losses are aggregated to update the policy.

\subsection{Rollout Collection}

Given a sample \(x\in\mathcal{D}\), the current policy \(\pi_\theta\) generates \(G\) responses:
\[
o_i \sim \pi_\theta(\cdot|x), \quad i=1,\dots,G.
\]
An outcome-level verifier evaluates each response:
\[
R(o_i,x)\in\{0,1\},
\]
where \(R(o_i,x)=1\) indicates that response \(o_i\) solves the task correctly.

The empirical rollout accuracy of sample \(x\) is
\[
a_x=
\frac{1}{G}
\sum_{i=1}^{G}
R(o_i,x).
\]
This quantity measures how often the current policy solves \(x\) under repeated sampling. Since rollouts are collected from the current policy, \(a_x\) changes as training proceeds.

\subsection{Behavior Signals}

We use two lightweight signals to describe the learning state of each sample: accuracy decomposition and confidence calibration. The resulting behavior state is
\[
b_x=(\omega(a_x),\varphi(c_x)).
\]

\paragraph{Accuracy decomposition.}
The scalar accuracy \(a_x\) is informative but coarse. A hard partition of \(a_x\) into low-, medium-, and high-accuracy regions would make routing unstable, especially when \(G\) is small and a single rollout can move a sample across a threshold. We therefore map \(a_x\) into three smooth membership scores:
\[
l_x=\exp\left(-\frac{(a_x-0)^2}{2\sigma_l^2}\right),
\]
\[
m_x=\exp\left(-\frac{(a_x-0.5)^2}{2\sigma_m^2}\right),
\]
\[
h_x=\exp\left(-\frac{(a_x-1)^2}{2\sigma_h^2}\right).
\]
After normalization,
\[
\omega(a_x)=[l,m,h]
=
\frac{[l_x,m_x,h_x]}{l_x+m_x+h_x}.
\]
Here \(l\), \(m\), and \(h\) correspond to low-accuracy, uncertain, and high-accuracy components. The centers \(0\), \(0.5\), and \(1\) match the natural interpretation of samples that the model mostly fails, solves inconsistently, or solves reliably. In our implementation, we fix
\[
\sigma_l=\sigma_h=0.18,\quad \sigma_m=0.16,
\]
and do not tune them per dataset.

\paragraph{Confidence calibration.}
For each response \(o_i=(y_{i,1},\dots,y_{i,T_i})\), we estimate the model's internal confidence from token-level predictive entropy. At position \(t\), let
\[
p_{\theta,t}(\cdot)=\pi_\theta(\cdot|x,y_{i,<t})
\]
be the next-token distribution. The token entropy is
\[
H_{i,t}
=
-\sum_{v\in\mathcal{V}}
p_{\theta,t}(v)\log p_{\theta,t}(v).
\]
The sequence-level uncertainty is
\[
\bar{H}(o_i,x)
=
\frac{1}{T_i}
\sum_{t=1}^{T_i}
H_{i,t}.
\]
We convert uncertainty into confidence using batch-level normalization:
\[
\mathrm{conf}(o_i,x)
=
1-\mathrm{Norm}_{\mathcal{B}}(\bar{H}(o_i,x)),
\]
where \(\mathrm{Norm}_{\mathcal{B}}\) maps entropy values from rollouts in the current batch to \([0,1]\). The sample-level confidence is
\[
c_x=
\frac{1}{G}
\sum_{i=1}^{G}
\mathrm{conf}(o_i,x).
\]

Since raw confidence can vary with prompt length, task type, and batch composition, we calibrate it within the current batch. Let
\[
\bar{c}_B=
\frac{1}{|\mathcal{B}|}
\sum_{x'\in\mathcal{B}}
c_{x'}.
\]
The relative confidence is
\[
\Delta c_x=c_x-\bar{c}_B.
\]
We normalize the confidence-related quantities and write
\[
\varphi(c_x)
=
(\widetilde{c},\widetilde{c}^{+},\widetilde{c}^{-},\widetilde{\Delta c}),
\]
where \(\widetilde{c}\) is normalized confidence, \(\widetilde{c}^{+}\) and \(\widetilde{c}^{-}\) indicate high- and low-confidence tendencies, and \(\widetilde{\Delta c}\) denotes the calibrated deviation from the batch mean.

\subsection{Recipe Router}

The router maps the behavior state \(b_x\) to a recipe distribution
\[
p_x=
[p_{\mathrm{GRPO}},p_{\mathrm{OPSD}},p_{\mathrm{REG}},p_{\mathrm{SKIP}}],
\]
with
\[
\sum_k p_{x,k}=1,\quad p_{x,k}\geq 0.
\]

To keep the router interpretable, we first compute four routing scores. Given
\[
\omega(a_x)=[l,m,h],
\]
we define
\[
s_{\mathrm{GRPO}}
=
m+h(1-\widetilde{\Delta c})+l(1-\widetilde{c}^{-})\cdot\widetilde{c},
\]
\[
s_{\mathrm{OPSD}}
=
l(1-\widetilde{c}^{-}),
\]
\[
s_{\mathrm{REG}}
=
h\cdot\widetilde{\Delta c}\cdot\widetilde{c}^{+},
\]
\[
s_{\mathrm{SKIP}}
=
l\cdot\widetilde{c}^{-}\cdot\widetilde{c}.
\]
These assignments are motivated by preliminary diagnostics on Qwen3-4B.
Mixed-correctness samples benefit more from GRPO, while low-accuracy,
low-confidence samples favor OPSD. Stable solved samples show limited
benefit from aggressive optimization, whereas confident failures benefit
from neither active recipe. These observations motivate routing the
corresponding states to GRPO, OPSD, REG, and SKIP, respectively.
Full diagnostics are reported in Appendix~\ref{app:router_diagnostics}.

The scores are normalized into probabilities:
\[
p_{x,k}
=
\frac{s_k}
{s_{\mathrm{GRPO}}+s_{\mathrm{OPSD}}+s_{\mathrm{REG}}+s_{\mathrm{SKIP}}+\epsilon}.
\]
Then each sample is assigned to one recipe:
\[
t_x\sim \mathrm{Categorical}(p_x).
\]
Thus, each sample contributes to at most one training objective in a given iteration.

\subsection{Recipe Assignment and Policy Update}

According to \(t_x\), the current batch is partitioned into four disjoint queues:
\[
\mathcal{B}_{g},\mathcal{B}_{o},\mathcal{B}_{r},\mathcal{B}_{s},
\]
corresponding to GRPO, OPSD, REG, and SKIP. Let their sizes be
\[
n_g=|\mathcal{B}_{g}|,\quad
n_o=|\mathcal{B}_{o}|,\quad
n_r=|\mathcal{B}_{r}|,\quad
n_s=|\mathcal{B}_{s}|.
\]

\paragraph{GRPO.}
For samples in \(\mathcal{B}_g\), we use the standard GRPO objective. For a rollout \(o_i\), let
\[
r_i = R(o_i,x).
\]
We compute the group mean and standard deviation as
\[
\mu_x = \frac{1}{G}\sum_{i=1}^{G} r_i,
\]
\[
\sigma_x =
\left[
\frac{1}{G}\sum_{i=1}^{G}
(r_i-\mu_x)^2
\right]^{1/2}.
\]
The group-relative advantage is
\[
A_i=\frac{r_i-\mu_x}{\sigma_x+\epsilon}.
\]
The policy ratio is
\[
\rho_i =
\frac{\pi_\theta(o_i|x)}
{\pi_{\theta_{\mathrm{old}}}(o_i|x)}.
\]
We use the clipped ratio
\[
\bar{\rho}_i =
\mathrm{clip}(\rho_i,1-\epsilon,1+\epsilon).
\]
The GRPO loss is
\[
\mathcal{L}_{\mathrm{GRPO}}
=
-\mathbb{E}
\left[
\min(\rho_i A_i,\bar{\rho}_i A_i)
\right],
\]
where the expectation is taken over \(x\in\mathcal{B}_g\) and rollout index \(i\).

\paragraph{OPSD.}
For samples in \(\mathcal{B}_o\), we use OPSD as a corrective learning recipe. The policy model acts as the student, while the same model conditioned on answer information acts as the teacher. Given the input \(x\), the student's current output, and the answer signal, the teacher produces a token-level target sequence
\[
y_x^T=(y_{x,1}^T,\dots,y_{x,T_x}^T).
\]
The student learns this target by token-level imitation:
\[
\mathcal{L}_{\mathrm{OPSD}}
=
-\frac{1}{|\mathcal{B}_{o}|}
\sum_{x\in\mathcal{B}_{o}}
\sum_{t=1}^{T_x}
\log
\pi_\theta
\left(
y_{x,t}^{T}
\mid
x,y_{x,<t}^{T}
\right).
\]
The teacher trajectory is generated once during offline preprocessing by
prompting the same base model with the problem and its ground-truth answer,
and is reused during training. Thus, OPSD requires no external teacher or
additional CoT annotation, but assumes access to target answers in the
verifiable post-training setting.

\paragraph{REG.}
For samples in \(\mathcal{B}_r\), we use a regularization objective. Since these samples already show reliable behavior, we constrain the updated policy to stay close to a reference policy:
\[
\mathcal{L}_{\mathrm{REG}}
=
\frac{1}{|\mathcal{B}_{r}|}
\sum_{x\in\mathcal{B}_{r}}
D_{\mathrm{KL}}
\left(
\pi_\theta(\cdot|x)
\|
\pi_{\mathrm{ref}}(\cdot|x)
\right).
\]
Here \(\pi_{\mathrm{ref}}\) can be the initial model, a pre-update checkpoint, or the old policy in the current iteration.

\paragraph{SKIP.}
Samples in \(\mathcal{B}_s\) are ignored in the current update:
\[
\mathcal{L}_{\mathrm{SKIP}}=0.
\]

The final training loss aggregates only active queues:
\[
\mathcal{L}
=
\frac{
n_g\mathcal{L}_{\mathrm{GRPO}}
+
n_o\mathcal{L}_{\mathrm{OPSD}}
+
n_r\mathcal{L}_{\mathrm{REG}}
}
{
n_g+n_o+n_r
}.
\]
SKIP samples do not contribute gradients and are excluded from the denominator.

\begin{table*}[t]
\centering
\small
\caption{Main results on ID mathematical reasoning and OOD general reasoning benchmarks. We report results across different Qwen3 and Qwen3.5 model scales. The best result in each column within the same model group is shown in \textbf{bold}, and the second-best result is \underline{underlined}.}
\label{tab:main_results}
\setlength{\tabcolsep}{4pt}
\begin{tabular}{llccccccc}
\toprule
\multirow{2}{*}{Model} & \multirow{2}{*}{Method}
& \multicolumn{4}{c}{ID Math Reasoning}
& \multicolumn{2}{c}{OOD General Reasoning}
& \multirow{2}{*}{Avg.} \\
\cmidrule(lr){3-6} \cmidrule(lr){7-8}
& & GSM8K & MATH-500 & AIME24 & AIME25 & MMLU-Pro & GPQA-diamond & \\
\midrule
\multirow{4}{*}{Qwen3-0.6B}
& Base & 65.8 & 35.7 & 11.6 & 8.7 & \textbf{23.4} & 20.6 & 27.6 \\
& Naive-GRPO & 70.4 & 42.8 & \underline{13.9} & 10.5 & 20.9 & \underline{22.1} & 30.1 \\
& Naive-OPSD & \underline{71.6} & \underline{44.1} & 13.5 & \underline{11.4} & 21.5 & 21.7 & \underline{30.6} \\
& Self-Routing & \textbf{73.2} & \textbf{45.8} & \textbf{15.6} & \textbf{13.0} & \underline{22.6} & \textbf{23.8} & \textbf{32.3} \\
\midrule
\multirow{4}{*}{Qwen3-1.7B}
& Base & 77.1 & 44.3 & 33.7 & 24.2 & \textbf{51.4} & 34.2 & 44.1 \\
& Naive-GRPO & 81.6 & 50.4 & 39.6 & 29.8 & 47.9 & 37.8 & 47.9 \\
& Naive-OPSD & \underline{83.4} & \underline{52.1} & \underline{42.7} & \underline{32.4} & 49.3 & \underline{39.1} & \underline{49.8} \\
& Self-Routing & \textbf{86.8} & \textbf{55.9} & \textbf{46.3} & \textbf{36.7} & \underline{50.6} & \textbf{42.5} & \textbf{53.1} \\
\midrule
\multirow{4}{*}{Qwen3-4B}
& Base & 82.3 & 54.4 & 66.9 & 54.1 & \textbf{59.2} & 49.3 & 61.0 \\
& Naive-GRPO & 89.6 & 62.8 & 74.3 & 65.4 & 54.6 & 53.8 & 66.8 \\
& Naive-OPSD & \underline{93.5} & \underline{66.1} & \underline{79.4} & \underline{70.2} & 56.4 & \underline{56.7} & \underline{70.4} \\
& Self-Routing & \textbf{95.2} & \textbf{69.8} & \textbf{83.6} & \textbf{74.9} & \underline{58.1} & \textbf{60.4} & \textbf{73.7} \\
\midrule
\multirow{4}{*}{Qwen3.5-0.8B}
& Base & 32.9 & 18.3 & 4.9 & 3.1 & \textbf{37.1} & 8.8 & 17.5 \\
& Naive-GRPO & 38.6 & 22.7 & \underline{6.8} & 4.7 & 33.8 & \underline{11.3} & 19.6 \\
& Naive-OPSD & \underline{39.8} & \underline{24.1} & 6.4 & \underline{5.5} & 34.7 & 10.9 & \underline{20.2} \\
& Self-Routing & \textbf{43.1} & \textbf{26.8} & \textbf{8.6} & \textbf{7.3} & \underline{36.0} & \textbf{13.6} & \textbf{22.6} \\
\midrule
\multirow{4}{*}{Qwen3.5-2B}
& Base & 69.1 & 44.2 & 24.1 & 17.3 & \textbf{57.4} & 31.5 & 40.6 \\
& Naive-GRPO & 77.9 & 50.7 & 36.8 & 28.1 & 53.8 & 36.9 & 47.4 \\
& Naive-OPSD & \underline{80.8} & \underline{54.3} & \underline{42.5} & \underline{32.6} & 55.1 & \underline{39.7} & \underline{50.8} \\
& Self-Routing & \textbf{84.6} & \textbf{58.2} & \textbf{48.4} & \textbf{38.5} & \underline{56.7} & \textbf{44.3} & \textbf{55.1} \\
\midrule
\multirow{4}{*}{Qwen3.5-4B}
& Base & 87.4 & 77.2 & 78.9 & 76.8 & \textbf{71.6} & 69.1 & 76.8 \\
& Naive-GRPO & 91.3 & 81.7 & 84.2 & 82.5 & 66.8 & 72.3 & 79.8 \\
& Naive-OPSD & \underline{93.4} & \underline{84.8} & \underline{88.6} & \underline{86.7} & 68.7 & \underline{75.9} & \underline{83.0} \\
& Self-Routing & \textbf{96.1} & \textbf{88.3} & \textbf{93.2} & \textbf{91.4} & \underline{70.9} & \textbf{79.6} & \textbf{86.6} \\
\bottomrule
\end{tabular}
\end{table*}

\section{Experiments}

We evaluate whether behavior-conditioned self-routing improves post-training over global recipes that apply the same objective to all samples. Our experiments answer four questions:

\begin{itemize}
    \item \textbf{Q1: Final performance.} Does behavior-conditioned routing improve reasoning performance compared with uniform post-training recipes and fixed recipe mixtures?
    \item \textbf{Q2: Efficiency.} Does routing reduce wasted updates by assigning expensive or unstable recipes only to samples where they are useful?
    \item \textbf{Q3: Routing strategy.} Is the proposed behavior-conditioned routing better than simpler alternatives, such as random routing, fixed-proportion routing, or hard threshold-based routing?
    \item \textbf{Q4: Training dynamics.} How does the recipe distribution change during training, and does it track the model's changing sample-level learning states?
\end{itemize}

\subsection{Experimental Setup}

\paragraph{Datasets.}
We use DAPO-Math-17K~\citep{yu2026dapo} as the training set for all post-training methods. For evaluation, we use four in-domain mathematical reasoning benchmarks and two out-of-domain general reasoning benchmarks. The in-domain benchmarks are GSM8K~\citep{cobbe2021training}, MATH-500~\citep{hendrycks2021math}, AIME24~\citep{maa2024aime}, and AIME25~\citep{maa2025aime}, covering grade-school arithmetic, competition-style mathematics, and recent exam-level problems. For out-of-domain evaluation, we use GPQA~\citep{rein2023gpqa} and MMLU-Pro~\citep{wang2024mmlu} to test whether math post-training transfers to broader reasoning tasks. We report the macro-average over all six evaluation benchmarks.

\paragraph{Models.}
We conduct experiments on Qwen3~\citep{yang2025qwen3} and Qwen3.5~\citep{team2026qwen3} models across small and medium scales. Specifically, we evaluate Qwen3-0.6B, Qwen3-1.7B, Qwen3-4B, Qwen3.5-0.8B, Qwen3.5-2B, and Qwen3.5-4B. This setting tests whether self-routing works across different model generations and base-model capabilities.

\paragraph{Baselines.}
For the main comparison, we include the base model without post-training, Naive-GRPO, and Naive-OPSD. Naive-GRPO applies GRPO uniformly to all training samples, while Naive-OPSD applies OPSD uniformly to all samples. These baselines test whether routing samples to different recipes improves over using a single global post-training recipe. Routing-specific baselines are discussed separately in Section~\ref{sec:routing_ablation}, where all variants use the same recipe set and differ only in the routing rule.
We additionally compare with DAPO-style RL and PODS~\citep{xu2025pods},
representing stabilized RL training and rollout selection, respectively;
these results are reported in Appendix~\ref{app:additional_results}.

\subsection{Main Results}
Table~\ref{tab:main_results} reports the main results on ID mathematical reasoning and OOD general reasoning benchmarks. Self-Routing achieves the highest average score across all evaluated backbones. The gains are most clear on ID math tasks, where Self-Routing consistently outperforms both Naive-GRPO and Naive-OPSD. For example, on Qwen3-4B, Self-Routing improves the average score from 61.0 to 73.7, exceeding Naive-GRPO and Naive-OPSD by 6.9 and 3.3 points, respectively. On Qwen3.5-4B, Self-Routing reaches 86.6 average score, compared with 79.8 for Naive-GRPO and 83.0 for Naive-OPSD.

The advantage of Self-Routing becomes larger as model capacity increases. On the smallest models, Naive-GRPO and Naive-OPSD still show mixed results on some individual benchmarks, such as AIME24 and GPQA-diamond, suggesting that weak models may not provide stable enough self-generated signals for OPSD to dominate in every case. For stronger backbones, OPSD becomes consistently better than GRPO, and Self-Routing further widens the margin. This trend supports our motivation that stronger models can benefit more from self-distillation and routing-based training.

For OOD general reasoning, we observe different behaviors on GPQA-diamond and MMLU-Pro. GPQA-diamond usually improves after math-oriented post-training, while MMLU-Pro drops compared with the base model. Although preserving MMLU-Pro performance is not the direct target of our method, Self-Routing shows the smallest degradation among all post-training methods and remains the second-best method after Base on every backbone. We conjecture that this smaller degradation may come from the conditional nature of Self-Routing: instead of applying a uniform math-oriented update to all training instances, it selects different self-improvement paths according to the model's own outputs. This may reduce the influence of noisy or overly specialized training signals, leading to less drift from the base model's original general-purpose behavior while still improving mathematical reasoning.

\paragraph{Stronger baselines and OOD transfer.}
On Qwen3-4B, Self-Routing also outperforms DAPO-style RL and PODS,
achieving 80.9/71.1/59.3 on ID math, OOD verifiable reasoning, and
general evaluation. On SATBench, AutoLogi, and LiveCodeBench-v5,
Self-Routing further obtains the best average score of 71.1.
Full results are provided in Appendix~\ref{app:additional_results}.

\subsection{Efficiency Analysis}
\label{sec:efficiency_analysis}

We provide a coarse-grained FLOPs estimate to analyze the training cost of different post-training recipes. Following the standard scaling-law approximation for Transformer language models, a forward pass costs approximately \(2N\) FLOPs per token, while a training pass with forward and backward computation costs approximately \(6N\) FLOPs per token~\citep{kaplan2020scaling,hoffmann2022training}. Let \(N\) denote the number of model parameters, \(T\) the number of training steps, \(B\) the batch size, \(G\) the number of rollouts per sample, and \(L\) the average sequence length. We normalize all costs by \(NTBL\) and ignore sequence-length variation and implementation overhead. The Self-Routing branch ratios are obtained by integrating the routing curves in Figure~\ref{fig:routing_dynamics}, giving \(30.8\%\) GRPO, \(30.4\%\) OPSD, \(25.5\%\) REG, and \(13.3\%\) SKIP.

As shown in Table~\ref{tab:flops-estimation}, Self-Routing is not intended as a wall-clock acceleration method under the current implementation. It is more expensive than Naive-OPSD because it retains multi-rollout behavior estimation and additionally applies GRPO-style updates to routed samples. However, it is less expensive than applying GRPO to all rollout groups. When \(G=8\), the normalized costs are \(64.0\) for Naive-GRPO, \(24.0\) for Naive-OPSD, and \(34.7\) for Self-Routing. The main efficiency benefit is therefore selective allocation: expensive optimization recipes are applied only to the subset of samples whose rollout behavior suggests that the corresponding training signal is useful.

\begin{table}[t]
\centering
\small
\begin{tabular}{lc}
\toprule
Method & Normalized FLOPs \\
\midrule
\textbf{Naive-GRPO} & \\
\quad Rollout & \(2G\) \\
\quad GRPO update & \(6G\) \\
\quad Total & \(8G\) \\
\midrule
\textbf{Naive-OPSD} & \\
\quad Rollout & \(2G\) \\
\quad Teacher generation & \(2\) \\
\quad SFT update & \(6\) \\
\quad Total & \(2G+8\) \\
\midrule
\textbf{Self-Routing} & \\
\quad Rollout & \(2G\) \\
\quad GRPO update & \(0.308 \times 6G\) \\
\quad OPSD update & \(0.304 \times (2+6)\) \\
\quad REG update & \(0.255 \times 6\) \\
\quad SKIP update & \(0\) \\
\quad Total & \(3.848G+3.962\) \\
\bottomrule
\end{tabular}
\caption{
Coarse-grained training FLOPs estimation normalized by \(NTBL\). We approximate inference as \(2N\) FLOPs per token and training as \(6N\) FLOPs per token. The Self-Routing ratios are obtained by integrating the routing curves in Figure~\ref{fig:routing_dynamics}.
}
\label{tab:flops-estimation}
\end{table}

\subsection{Ablation on Routing Strategies}
\label{sec:routing_ablation}

\begin{figure}[t]
    \centering
    \includegraphics[width=\linewidth]{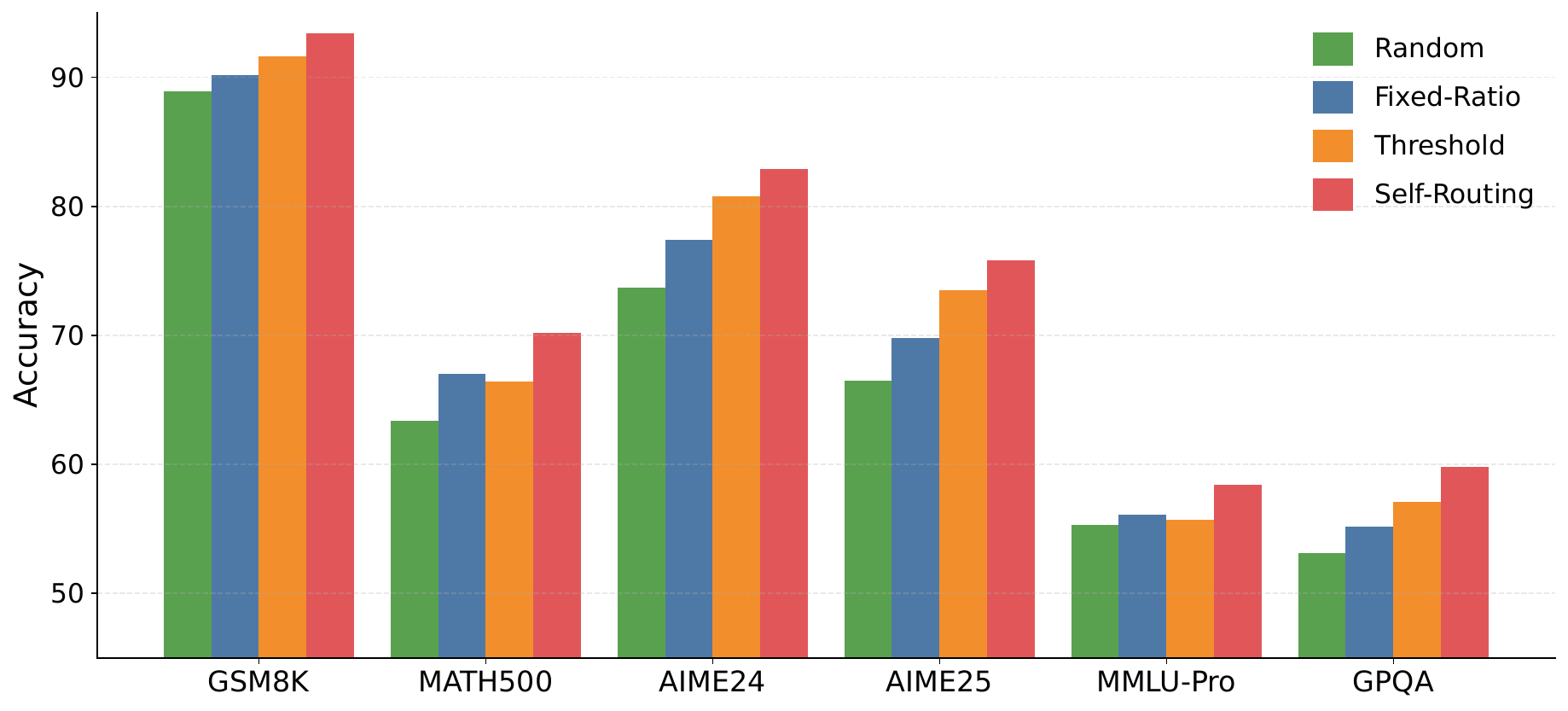}
    \caption{Ablation on routing strategies using Qwen3-4B. We compare Self-Routing with three simpler routing baselines across ID mathematical reasoning and OOD general reasoning benchmarks.}
    \label{fig:routing_ablation}
\end{figure}

To verify the effectiveness of Self-Routing, we compare it with three simpler routing strategies on Qwen3-4B. Round-wise Random randomly selects one recipe for each training round and applies it to all samples in that round. Fixed-Ratio Random assigns samples within each round according to a fixed Reg:GRPO:OPSD:Skip ratio of 3:3:3:1. Accuracy-Based Routing first sorts samples by their current accuracy, then routes the lowest 30\% to Reg, the next 30\% to GRPO, the next 30\% to OPSD, and the highest 10\% to Skip. These baselines form a progression from recipe-level randomness, to fixed recipe mixing, and then to a simple sample-level routing rule.

As shown in Figure~\ref{fig:routing_ablation}, the two random routing baselines perform poorly, suggesting that simply alternating or mixing recipes is not enough. Accuracy-Based Routing gives a much stronger result and even outperforms Naive-OPSD in the main table, which shows that sample-level routing signals are important. However, it still falls slightly behind Self-Routing across most benchmarks. This gap indicates that accuracy alone is not sufficient for recipe assignment: samples with the same correctness can differ in reasoning quality, confidence, stability, and optimization risk. By using richer behavior-conditioned signals, Self-Routing makes finer routing decisions and achieves the best final performance.

More targeted ablations further show that accuracy is the primary routing
signal, while confidence calibration and the behavior-conditioned
assignment provide complementary gains; see Appendix~\ref{app:router_ablation}.

\subsection{Training Dynamics}
\label{sec:training_dynamics}

\begin{figure}[t]
    \centering
    \includegraphics[width=\linewidth]{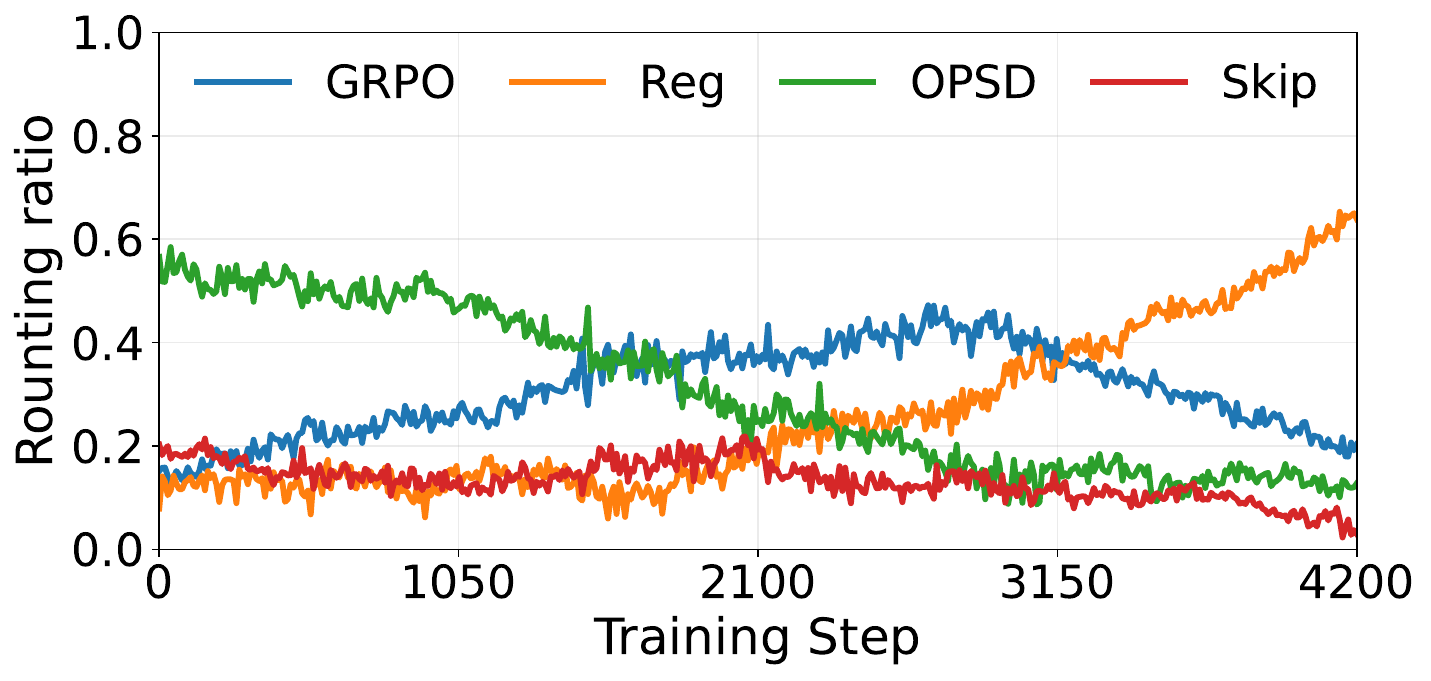}
    \caption{Routing ratios of different training recipes during Self-Routing training.}
    \label{fig:routing_dynamics}
\end{figure}

Figure~\ref{fig:routing_dynamics} illustrates the routing dynamics of Self-Routing on Qwen3-4B during training. In practice, we sample the routing decisions every ten training steps and aggregate them to obtain the counting ratio of each branch, which smooths out short-term fluctuations while preserving the overall trend. The figure shows that the routing distribution changes substantially as training proceeds, rather than staying fixed around a preset ratio.

At the early stage, OPSD accounts for the largest proportion, with a counting ratio above 0.5, while GRPO and Reg remain relatively low. As training continues, OPSD gradually decreases, while GRPO increases and reaches its peak in the middle stage, suggesting that more samples become suitable for reward-driven optimization after the model has acquired stronger reasoning behavior. In the later stage, Reg rises quickly and becomes the dominant branch, while both OPSD and GRPO decline; this may help constrain further policy drift after sufficient reasoning improvement has been obtained. The Skip branch remains relatively low throughout training and further decreases near the end, indicating that most samples can still provide useful training signals through one of the active branches.

\section{Conclusion}

We presented Self-Routing, a self-contained post-training framework that turns the model's own rollout behavior into sample-level training recipes. Instead of applying one objective to all data, Self-Routing uses rollout correctness and confidence to route each sample to GRPO, on-policy self-distillation, regularization, or skipping. Experiments on mathematical reasoning show consistent gains over uniform GRPO, uniform OPSD, fixed mixtures, and simpler routing baselines across Qwen3 and Qwen3.5 backbones. The routing dynamics further show that different recipes become useful at different training stages. These results suggest that rollout behavior should guide not only which samples are used, but also how they are optimized.

\section*{Limitations}

Although our method demonstrates promising empirical improvements, several limitations remain. Our current study mainly focuses on the intuition that heterogeneous rollout states may benefit from different post-training strategies, and we validate this idea through empirical performance gains and behavioral observations. However, the work does not yet provide a sufficiently deep mechanistic or theoretical explanation for why certain rollout patterns align better with specific optimization signals. At present, the routing behavior is primarily supported by intuition and experimental evidence rather than a rigorous understanding of the underlying optimization dynamics, which remains an important direction for future research.

In addition, our training experiments are centered on mathematical
reasoning. Although we observe transfer to SATBench, AutoLogi, and
LiveCodeBench-v5, these benchmarks still primarily cover related
verifiable reasoning tasks. It remains unclear whether the same routing paradigm generalizes to substantially different settings such as agent planning or open-ended instruction following.

Finally, it is important to note that our work is centered on the routing perspective itself rather than the optimization of individual training algorithms. We do not attempt to improve the internal designs of specific methods, such as reward engineering for GRPO-style reinforcement learning or trajectory/path refinement strategies for OPSD-like approaches. These algorithmic improvements are largely orthogonal to our objective. Instead, our focus is on exploring whether different optimization mechanisms should be assigned adaptively according to rollout states under a unified post-training framework.

\section*{Ethical Considerations}

In this work, we use AI-assisted tools to polish the writing and improve code quality. We plan to release our code and data to facilitate future research and reproducibility. For all environments, datasets, models, and external resources used in this work, we strictly follow their original usage terms and licensing agreements, and confirm that all artifacts are used solely for academic research purposes. In addition, several icons used in our figures are obtained from the FLATICON website.

\section*{Acknowledgments}

This work was supported by Fundamental and Interdisciplinary Disciplines Breakthrough Plan of the Ministry of Education of China (JYB2025XDXM116), National Natural Science Foundation of China (No. 62137002, 62293550, 62293553, 62293554, 62437002, 62477036, 62477037, 62192781), the Shaanxi Provincial Social Science Foundation Project (No. 2024P041), the Youth Innovation Team of Shaanxi Universities "Multi-modal Data Mining and Fusion", and Xi'an Jiaotong University City College Research Project (No. 2024Y01), and the Zhongguancun Academy (Grant No. 20240103).

\bibliography{latex/custom}

@article{yang2025qwen3,
  title={Qwen3 technical report},
  author={Yang, An and Li, Anfeng and Yang, Baosong and Zhang, Beichen and Hui, Binyuan and Zheng, Bo and Yu, Bowen and Gao, Chang and Huang, Chengen and Lv, Chenxu and others},
  journal={arXiv preprint arXiv:2505.09388},
  year={2025}
}

@article{kaplan2020scaling,
  title={Scaling laws for neural language models},
  author={Kaplan, Jared and McCandlish, Sam and Henighan, Tom and Brown, Tom B and Chess, Benjamin and Child, Rewon and Gray, Scott and Radford, Alec and Wu, Jeffrey and Amodei, Dario},
  journal={arXiv preprint arXiv:2001.08361},
  year={2020}
}

@article{hoffmann2022training,
  title={Training compute-optimal large language models},
  author={Hoffmann, Jordan and Borgeaud, Sebastian and Mensch, Arthur and Buchatskaya, Elena and Cai, Trevor and Rutherford, Eliza and Casas, DDL and Hendricks, Lisa Anne and Welbl, Johannes and Clark, Aidan and others},
  journal={arXiv preprint arXiv:2203.15556},
  volume={10},
  year={2022}
}

@article{rein2023gpqa,
  title={Gpqa: A graduate-level google-proof q\&a benchmark},
  author={Rein, David and Hou, Betty Li and Stickland, Asa Cooper and Petty, Jackson and Pang, Richard Yuanzhe and Dirani, Julien and Michael, Julian and Bowman, Samuel R},
  journal={arXiv preprint arXiv:2311.12022},
  year={2023}
}

@article{wang2024mmlu,
  title={Mmlu-pro: A more robust and challenging multi-task language understanding benchmark},
  author={Wang, Yubo and Ma, Xueguang and Zhang, Ge and Ni, Yuansheng and Chandra, Abhranil and Guo, Shiguang and Ren, Weiming and Arulraj, Aaran and He, Xuan and Jiang, Ziyan and others},
  journal={Advances in Neural Information Processing Systems},
  volume={37},
  pages={95266--95290},
  year={2024}
}

@article{yu2026dapo,
  title={Dapo: An open-source llm reinforcement learning system at scale},
  author={Yu, Qiying and Zhang, Zheng and Zhu, Ruofei and Yuan, Yufeng and Zuo, Xiaochen and Yue, Yu and Dai, Weinan and Fan, Tiantian and Liu, Gaohong and Liu, Lingjun and others},
  journal={Advances in Neural Information Processing Systems},
  volume={38},
  pages={113222--113244},
  year={2026}
}

@article{shao2024deepseekmath,
  title={Deepseekmath: Pushing the limits of mathematical reasoning in open language models},
  author={Shao, Zhihong and Wang, Peiyi and Zhu, Qihao and Xu, Runxin and Song, Junxiao and Bi, Xiao and Zhang, Haowei and Zhang, Mingchuan and Li, YK and Wu, Yang and others},
  journal={arXiv preprint arXiv:2402.03300},
  year={2024}
}

@inproceedings{tan2023gkd,
  title={Gkd: A general knowledge distillation framework for large-scale pre-trained language model},
  author={Tan, Shicheng and Tam, Weng Lam and Wang, Yuanchun and Gong, Wenwen and Zhao, Shu and Zhang, Peng and Tang, Jie},
  booktitle={Proceedings of the 61st Annual Meeting of the Association for Computational Linguistics (Volume 5: Industry Track)},
  pages={134--148},
  year={2023}
}

@article{zhao2026self,
  title={Self-Distilled Reasoner: On-Policy Self-Distillation for Large Language Models},
  author={Zhao, Siyan and Xie, Zhihui and Liu, Mengchen and Huang, Jing and Pang, Guan and Chen, Feiyu and Grover, Aditya},
  journal={arXiv preprint arXiv:2601.18734},
  year={2026}
}

@article{team2026qwen3,
  title={Qwen3.5-omni technical report},
  author={Team, Qwen},
  journal={arXiv preprint arXiv:2604.15804},
  year={2026}
}

@inproceedings{zhao2025swift,
  title={Swift: a scalable lightweight infrastructure for fine-tuning},
  author={Zhao, Yuze and Huang, Jintao and Hu, Jinghan and Wang, Xingjun and Mao, Yunlin and Zhang, Daoze and Jiang, Zeyinzi and Wu, Zhikai and Ai, Baole and Wang, Ang and others},
  booktitle={Proceedings of the AAAI Conference on Artificial Intelligence},
  volume={39},
  pages={29733--29735},
  year={2025}
}

@article{paszke2019pytorch,
  title={Pytorch: An imperative style, high-performance deep learning library},
  author={Paszke, Adam and Gross, Sam and Massa, Francisco and Lerer, Adam and Bradbury, James and Chanan, Gregory and Killeen, Trevor and Lin, Zeming and Gimelshein, Natalia and Antiga, Luca and others},
  journal={Advances in neural information processing systems},
  volume={32},
  year={2019}
}

@article{guo2025deepseekr1,
  title={{DeepSeek-R1}: Incentivizing Reasoning Capability in {LLMs} via Reinforcement Learning},
  author={{DeepSeek-AI} and Guo, Daya and Yang, Dejian and Zhang, Haowei and Song, Junxiao and Wang, Peiyi and Zhu, Qihao and Xu, Runxin and Zhang, Ruoyu and Ma, Shirong and others},
  journal={arXiv preprint arXiv:2501.12948},
  year={2025}
}

@article{cobbe2021training,
  title={Training Verifiers to Solve Math Word Problems},
  author={Cobbe, Karl and Kosaraju, Vineet and Bavarian, Mohammad and Chen, Mark and Jun, Heewoo and Kaiser, Lukasz and Plappert, Matthias and Tworek, Jerry and Hilton, Jacob and Nakano, Reiichiro and Hesse, Christopher and Schulman, John},
  journal={arXiv preprint arXiv:2110.14168},
  year={2021}
}

@article{uesato2022solving,
  title={Solving Math Word Problems with Process- and Outcome-Based Feedback},
  author={Uesato, Jonathan and Kushman, Nate and Kumar, Ramana and Song, Francis and Siegel, Noah and Wang, Lisa and Creswell, Antonia and Irving, Geoffrey and Higgins, Irina},
  journal={arXiv preprint arXiv:2211.14275},
  year={2022}
}

@article{le2022coderl,
  title={{CodeRL}: Mastering Code Generation through Pretrained Models and Deep Reinforcement Learning},
  author={Le, Hung and Wang, Yue and Gotmare, Akhilesh Deepak and Savarese, Silvio and Hoi, Steven C. H.},
  journal={arXiv preprint arXiv:2207.01780},
  year={2022}
}

@article{liu2023rltf,
  title={{RLTF}: Reinforcement Learning from Unit Test Feedback},
  author={Liu, Jiate and Zhu, Yiqin and Xiao, Kaiwen and Fu, Qiang and Han, Xiao and Yang, Wei and Ye, Deheng},
  journal={arXiv preprint arXiv:2307.04349},
  year={2023}
}

@article{gehring2024rlef,
  title={{RLEF}: Grounding Code {LLMs} in Execution Feedback with Reinforcement Learning},
  author={Gehring, Jonas and Zheng, Kunhao and Copet, Jade and Mella, Vegard and Carbonneaux, Quentin and Cohen, Taco and Synnaeve, Gabriel},
  journal={arXiv preprint arXiv:2410.02089},
  year={2024}
}

@article{wei2025swerl,
  title={{SWE-RL}: Advancing {LLM} Reasoning via Reinforcement Learning on Open Software Evolution},
  author={Wei, Yuxiang and Duchenne, Olivier and Copet, Jade and Carbonneaux, Quentin and Zhang, Lingming and Fried, Daniel and Synnaeve, Gabriel and Singh, Rishabh and Wang, Sida I.},
  journal={arXiv preprint arXiv:2502.18449},
  year={2025}
}

@article{wang2022selfconsistency,
  title={Self-Consistency Improves Chain of Thought Reasoning in Language Models},
  author={Wang, Xuezhi and Wei, Jason and Schuurmans, Dale and Le, Quoc and Chi, Ed and Narang, Sharan and Chowdhery, Aakanksha and Zhou, Denny},
  journal={arXiv preprint arXiv:2203.11171},
  year={2022}
}

@article{weng2022selfverification,
  title={Large Language Models are Better Reasoners with Self-Verification},
  author={Weng, Yixuan and Zhu, Minjun and Xia, Fei and Li, Bin and He, Shizhu and Liu, Shengping and Sun, Bin and Liu, Kang and Zhao, Jun},
  journal={arXiv preprint arXiv:2212.09561},
  year={2022}
}

@article{agarwal2023onpolicy,
  title={On-Policy Distillation of Language Models: Learning from Self-Generated Mistakes},
  author={Agarwal, Rishabh and Vieillard, Nino and Zhou, Yongchao and Stanczyk, Piotr and Ramos, Sabela and Geist, Matthieu and Bachem, Olivier},
  journal={arXiv preprint arXiv:2306.13649},
  year={2023}
}

@article{schulman2017ppo,
  title={Proximal Policy Optimization Algorithms},
  author={Schulman, John and Wolski, Filip and Dhariwal, Prafulla and Radford, Alec and Klimov, Oleg},
  journal={arXiv preprint arXiv:1707.06347},
  year={2017}
}

@article{kimi2025k1,
  title={{Kimi k1.5}: Scaling Reinforcement Learning with {LLMs}},
  author={{Kimi Team}},
  journal={arXiv preprint arXiv:2501.12599},
  year={2025}
}

@article{hu2025openreasoner,
  title={Open-Reasoner-Zero: An Open Source Approach to Scaling Up Reinforcement Learning on the Base Model},
  author={Hu, Jingcheng and Zhang, Yinmin and Han, Qi and Jiang, Daxin and Zhang, Xiangyu and Shum, Heung-Yeung},
  journal={arXiv preprint arXiv:2503.24290},
  year={2025}
}

@article{zhao2025absolutezero,
  title={Absolute Zero: Reinforced Self-play Reasoning with Zero Data},
  author={Zhao, Andrew and Wu, Yiran and Yue, Yang and Wu, Tong and Xu, Quentin and Yue, Yang and Lin, Matthieu and Wang, Shenzhi and Wu, Qingyun and Zheng, Zilong and Huang, Gao},
  journal={arXiv preprint arXiv:2505.03335},
  year={2025}
}

@article{wang2025entropytokens,
  title={Beyond the 80/20 Rule: High-Entropy Minority Tokens Drive Effective Reinforcement Learning for {LLM} Reasoning},
  author={Wang, Shenzhi and Yu, Le and Gao, Chang and Zheng, Chujie and Liu, Shixuan and Lu, Rui and Dang, Kai and Chen, Xionghui and Yang, Jianxin and Zhang, Zhenru and Liu, Yuqiong and Yang, An and Zhao, Andrew and Yue, Yang and Song, Shiji and Yu, Bowen and Huang, Gao and Lin, Junyang},
  journal={arXiv preprint arXiv:2506.01939},
  year={2025}
}

@article{xu2025pods,
  title={Not All Rollouts are Useful: Down-Sampling Rollouts in {LLM} Reinforcement Learning},
  author={Xu, Yixuan Even and Savani, Yash and Fang, Fei and Kolter, J. Zico},
  journal={arXiv preprint arXiv:2504.13818},
  year={2025}
}

@article{zelikman2022star,
  title={{STaR}: Bootstrapping Reasoning With Reasoning},
  author={Zelikman, Eric and Wu, Yuhuai and Mu, Jesse and Goodman, Noah D.},
  journal={arXiv preprint arXiv:2203.14465},
  year={2022}
}

@article{zhang2025bread,
  title={{BREAD}: Branched Rollouts from Expert Anchors Bridge {SFT} and {RL} for Reasoning},
  author={Zhang, Xuechen and Huang, Zijian and Li, Yingcong and Ni, Chenshun and Chen, Jiasi and Oymak, Samet},
  journal={arXiv preprint arXiv:2506.17211},
  year={2025}
}

@article{hubotter2026selfdistillation,
  title={Reinforcement Learning via Self-Distillation},
  author={H{\"u}botter, Jonas and L{\"u}beck, Frederike and Behric, Lejs and Baumann, Anton and Bagatella, Marco and Marta, Daniel and Hakimi, Ido and Shenfeld, Idan and Kleine Buening, Thomas and Guestrin, Carlos and Krause, Andreas},
  journal={arXiv preprint arXiv:2601.20802},
  year={2026}
}

@article{wang2025oneshotrlvr,
  title={Reinforcement Learning for Reasoning in Large Language Models with One Training Example},
  author={Wang, Yiping and Yang, Qing and Zeng, Zhiyuan and Ren, Liliang and Liu, Lucas and Peng, Baolin and Cheng, Hao and He, Xuehai and Wang, Kuan and Gao, Jianfeng and Chen, Weizhu and Wang, Shuohang and Du, Simon Shaolei and Shen, Yelong},
  journal={arXiv preprint arXiv:2504.20571},
  year={2025}
}

@article{jiang2025vcrl,
  title={{VCRL}: Variance-based Curriculum Reinforcement Learning for Large Language Models},
  author={Jiang, Guochao and Feng, Wenfeng and Quan, Guofeng and Hao, Chuzhan and Zhang, Yuewei and Liu, Guohua and Wang, Hao},
  journal={arXiv preprint arXiv:2509.19803},
  year={2025}
}

@article{chen2023universal,
  title={Universal Self-Consistency for Large Language Model Generation},
  author={Chen, Xinyun and Aksitov, Renat and Alon, Uri and Ren, Jie and Xiao, Kefan and Yin, Pengcheng and Prakash, Sushant and Sutton, Charles and Wang, Xuezhi and Zhou, Denny},
  journal={arXiv preprint arXiv:2311.17311},
  year={2023}
}

@article{yao2023tree,
  title={Tree of Thoughts: Deliberate Problem Solving with Large Language Models},
  author={Yao, Shunyu and Yu, Dian and Zhao, Jeffrey and Shafran, Izhak and Griffiths, Thomas L. and Cao, Yuan and Narasimhan, Karthik},
  journal={arXiv preprint arXiv:2305.10601},
  year={2023}
}

@inproceedings{jiang2023llmblender,
  title={{LLM-Blender}: Ensembling Large Language Models with Pairwise Ranking and Generative Fusion},
  author={Jiang, Dongfu and Ren, Xiang and Lin, Bill Yuchen},
  booktitle={Proceedings of the 61st Annual Meeting of the Association for Computational Linguistics},
  year={2023}
}

@article{zuo2025ttrl,
  title={{TTRL}: Test-Time Reinforcement Learning},
  author={Zuo, Yuxin and Zhang, Kaiyan and Sheng, Li and Qu, Shang and Cui, Ganqu and Zhu, Xuekai and Li, Haozhan and Zhang, Yuchen and Long, Xinwei and Hua, Ermo and Qi, Biqing and Sun, Youbang and Ma, Zhiyuan and Yuan, Lifan and Ding, Ning and Zhou, Bowen},
  journal={arXiv preprint arXiv:2504.16084},
  year={2025}
}

@article{zhang2025corewarding,
  title={Co-rewarding: Stable Self-supervised Reinforcement Learning for Eliciting Reasoning in Large Language Models},
  author={Zhang, Zizhuo and Zhu, Jianing and Ge, Xinmu and Zhao, Zihua and Zhou, Zhanke and Li, Xuan and Feng, Xiao and Yao, Jiangchao and Han, Bo},
  journal={arXiv preprint arXiv:2508.00410},
  year={2025}
}

@article{hendrycks2021math,
  title={Measuring Mathematical Problem Solving With the {MATH} Dataset},
  author={Hendrycks, Dan and Burns, Collin and Kadavath, Saurav and Arora, Akul and Basart, Steven and Tang, Eric and Song, Dawn and Steinhardt, Jacob},
  journal={arXiv preprint arXiv:2103.03874},
  year={2021}
}

@misc{maa2024aime,
  title={2024 American Invitational Mathematics Examination},
  author={{Mathematical Association of America}},
  year={2024},
  howpublished={American Mathematics Competitions},
  url={https://maa.org/math-competitions/american-invitational-mathematics-examination-aime}
}

@misc{maa2025aime,
  title={2025 American Invitational Mathematics Examination},
  author={{Mathematical Association of America}},
  year={2025},
  howpublished={American Mathematics Competitions},
  url={https://maa.org/math-competitions/american-invitational-mathematics-examination-aime}
}

@article{ma2025brittle,
  title={How Brittle is Agent Safety? Rethinking Agent Risk under Intent Concealment and Task Complexity},
  author={Ma, Zihan and Zhu, Dongsheng and Liu, Shudong and Zhang, Taolin and Liu, Junnan and Li, Qingqiu and Luo, Minnan and Zhang, Songyang and Chen, Kai},
  journal={arXiv preprint arXiv:2511.08487},
  year={2025}
}

@article{xu2026dual,
  title={Dual-Dimensional Consistency: Balancing Budget and Quality in Adaptive Inference-Time Scaling},
  author={Xu, Rongman and Li, Yifei and Zhao, Tianzhe and Wu, Yanrui and Li, Bo and Yan, Hang},
  journal={arXiv preprint arXiv:2605.15100},
  year={2026}
}

@article{yan2026maga,
  title={MAGA: Multi-Platform Self-Fusion of GUI Agents via Structured Action Distillation},
  author={Yan, Hang and Gu, Zhangxuan and Zhou, Beitong and Chen, Jiaxuan and Li, Runze and Hu, Yusong and Shen, Shuheng and Meng, Changhua},
  journal={arXiv preprint arXiv:2607.29320},
  year={2026}
}

@inproceedings{yan2026mur,
  title={Mur: Momentum uncertainty guided reasoning for large language models},
  author={Yan, Hang and Xu, Fangzhi and Xu, Rongman and Li, Yifei and Zhang, Jian and Luo, Haoran and Wu, Xiaobao and Tuan, Luu Anh and Zhao, Haiteng and Lin, Qika and others},
  booktitle={Proceedings of the 64th Annual Meeting of the Association for Computational Linguistics (Volume 1: Long Papers)},
  pages={23078--23103},
  year={2026}
}

@inproceedings{li2026locomo,
  title={Locomo-plus: Beyond-factual cognitive memory evaluation framework for llm agents},
  author={Li, Yifei and Guo, Weidong and Zhang, Lingling and Xu, Rongman and Huang, Muye and Liu, Hui and Xu, Lijiao and Xu, Yu and Liu, Jun},
  booktitle={Proceedings of the 64th Annual Meeting of the Association for Computational Linguistics (Volume 1: Long Papers)},
  pages={25085--25100},
  year={2026}
}

@misc{li2026selfspecializedteachersdomainposttraining,
  title        = {Self-Specialized Teachers for Domain Post-Training},
  author       = {Yifei Li and Rongman Xu and Lingling Zhang and Muye Huang
                  and Zihan Ma and Jiashuai Liu and Hang Yan and Heng Wang},
  year         = {2026},
  eprint       = {2608.28647},
  archivePrefix = {arXiv},
  primaryClass = {cs.AI},
  url          = {https://arxiv.org/abs/2608.28647}
}


\appendix

\section{Self-Routing Algorithm}

Algorithm~\ref{alg:self-routing} gives the full training procedure. All recipes share the same sampled batch and the same on-policy rollouts. The router only changes how each sample contributes to the update.

\begin{algorithm}[t]
\caption{Self-Routing Post-Training}
\label{alg:self-routing}
\begin{algorithmic}[1]
\Require Training set $\mathcal{D}$, policy $\pi_\theta$, verifier $R$, rollout count $G$, reference policy $\pi_{\mathrm{ref}}$
\For{training step $t=1,\ldots,T$}
    \State Sample a mini-batch $B \subset \mathcal{D}$
    \For{each sample $x \in B$}
        \State Generate on-policy rollouts $\{o_i\}_{i=1}^{G}$ from $\pi_\theta(\cdot|x)$
        \State Evaluate each rollout with the verifier: $r_i = R(o_i,x)$
        \State Compute rollout accuracy $a_x = \frac{1}{G}\sum_{i=1}^{G} r_i$
        \State Estimate sample confidence $c_x$ from token-level predictive entropy
        \State Build behavior state $b_x=(\omega(a_x),\phi(c_x))$
        \State Compute routing scores $s_{\mathrm{GRPO}},s_{\mathrm{OPSD}},s_{\mathrm{REG}},s_{\mathrm{SKIP}}$
        \State Normalize scores into $p_x=[p_{\mathrm{GRPO}},p_{\mathrm{OPSD}},p_{\mathrm{REG}},p_{\mathrm{SKIP}}]$
        \State Assign recipe $z_x \sim \mathrm{Categorical}(p_x)$
    \EndFor
    \State Partition $B$ into $B_g,B_o,B_r,B_s$ according to recipe assignments
    \State Compute $\mathcal{L}_{\mathrm{GRPO}}$ on $B_g$
    \State Compute $\mathcal{L}_{\mathrm{OPSD}}$ on $B_o$
    \State Compute $\mathcal{L}_{\mathrm{REG}}$ on $B_r$
    \State Skip samples in $B_s$
    \State Update $\pi_\theta$ with
    \[
    \mathcal{L}=
    \frac{
    |B_g|\mathcal{L}_{\mathrm{GRPO}}
    + |B_o|\mathcal{L}_{\mathrm{OPSD}}
    + |B_r|\mathcal{L}_{\mathrm{REG}}
    }{|B_g|+|B_o|+|B_r|}.
    \]
\EndFor
\end{algorithmic}
\end{algorithm}

\section{Implementation Details}

We implement the training pipeline with \texttt{ms-swift}~\citep{zhao2025swift}. The framework handles model loading, distributed training, rollout generation, and optimizer execution. We add the routing module as a lightweight layer between rollout collection and loss construction. This keeps the baseline recipes and Self-Routing under the same training backend: the same data loader, rollout interface, verifier calls, tokenizer, and checkpointing code are used across Naive-GRPO, Naive-OPSD, and Self-Routing.

For each sampled prompt, the current policy first generates $G$ responses. The verifier returns a binary outcome reward for each response. The routing module then computes two signals: empirical rollout accuracy and entropy-based confidence. The router does not call an external teacher model, reward model, or data filter. OPSD targets are generated once during offline preprocessing by conditioning the same base model on the ground-truth answer, and are reused throughout training. GRPO and REG use the on-policy rollouts and reference policy already available in the training job.

\paragraph{Backend and execution.}
All experiments use \texttt{ms-swift} as the post-training launcher. The routing code is implemented as a recipe assignment module that writes four disjoint queues in each batch: GRPO, OPSD, REG, and SKIP. The update step reads these queues and applies the corresponding loss. This design also makes the ablation baselines easy to run, since random routing, fixed-ratio routing, and accuracy-only routing only replace the assignment rule.

\paragraph{Runtime environment.}
Training is conducted on a single node with 8 NVIDIA A100 GPUs. The codebase is built on PyTorch~\citep{paszke2019pytorch} and uses the standard Hugging Face model/tokenizer interface through \texttt{ms-swift}. Multi-GPU execution, mixed-precision training, checkpoint saving, and distributed data loading are handled by the training backend. We keep the same runtime stack for all compared methods, so differences in the reported results do not come from different launchers or inference engines.

\paragraph{Reference policy.}
For the REG branch, we use a fixed reference policy during each update. In practice, this can be the initial model, the old policy before the current update, or a checkpoint selected by the training script. In our runs, the reference is passed through the same \texttt{ms-swift} model interface as the policy model, so no separate inference service is needed.

\paragraph{Verifier.}
For mathematical reasoning tasks, the verifier extracts the final answer and compares it with the ground-truth answer. We use the same verifier for rollout scoring, GRPO rewards, and evaluation-time correctness. This avoids giving Self-Routing a stronger correctness signal than the baselines.

\section{Routing Hyperparameters}

We keep the routing-related hyperparameters fixed across datasets and model scales. Each prompt uses $G=8$ on-policy rollouts. The verifier reward is binary correctness, $R(o,x)\in\{0,1\}$. The accuracy membership centers are $0$, $0.5$, and $1$, corresponding to low-accuracy, uncertain, and high-accuracy behavior states. We set $\sigma_l=\sigma_h=0.18$ and $\sigma_m=0.16$.

The confidence signal is the mean token-level predictive entropy of the generated response, normalized within the current training batch. The active recipe set contains GRPO, OPSD, and REG, while SKIP contributes no gradient in the current update. Recipe assignment uses categorical sampling from the normalized routing scores.

\section{Training and Evaluation Configuration}

We use DAPO-Math-17K as the training set and evaluate on GSM8K, MATH-500, AIME24, AIME25, MMLU-Pro, and GPQA-diamond. The reported average is the macro-average over these six benchmarks. The main baselines are Base, Naive-GRPO, and Naive-OPSD. The routing baselines are round-wise random routing, fixed-ratio random routing, and accuracy-based routing.

All training runs use the same \texttt{ms-swift} post-training framework. Experiments are conducted on a single node with 8 NVIDIA A100 GPUs. The implementation is based on PyTorch and uses Hugging Face Transformers-compatible model and tokenizer APIs through \texttt{ms-swift}. The same runtime stack is used for Naive-GRPO, Naive-OPSD, routing baselines, and Self-Routing. No external reward model or extra annotation is used.

\section{Recipe Assignment Rules}

Each sample is then assigned to one recipe by categorical sampling. We use sampling rather than an argmax rule because the behavior estimates come from a finite rollout group. Sampling also prevents a large number of borderline samples from collapsing into the same branch early in training.

The branch meanings are as follows. GRPO handles samples whose rollouts contain enough variation to produce a useful relative reward signal. OPSD handles failed but recoverable samples, where dense imitation can give a stronger update than sparse outcome reward. REG handles samples that the model already solves with high confidence. SKIP removes samples whose current behavior provides little training signal for the active recipes.

\section{FLOPs Estimate}

We use the standard Transformer cost approximation: a forward pass costs about $2N$ FLOPs per token and a training pass with forward and backward computation costs about $6N$ FLOPs per token, where $N$ is the parameter count. Let $T$ denote training steps, $B$ batch size, $G$ rollouts per sample, and $L$ average sequence length. After normalizing by $NTBL$, the main costs are:
\[
\mathrm{Naive\text{-}GRPO}=2G+6G=8G,
\]
\[
\mathrm{Naive\text{-}OPSD}=2G+2+6=2G+8.
\]
For Self-Routing, the measured average branch ratios are $30.8\%$ GRPO, $30.4\%$ OPSD, $25.5\%$ REG, and $13.3\%$ SKIP. Its normalized cost is therefore

\[
\begin{aligned}
&2G + 0.308\cdot 6G + 0.304\cdot(2+6)
   + 0.255\cdot 6 \\
&\qquad = 3.848G + 3.962.
\end{aligned}
\]
With $G=8$, the normalized costs are $64.0$ for Naive-GRPO, $24.0$ for Naive-OPSD, and $34.7$ for Self-Routing. Self-Routing is not meant to be the cheapest recipe in this implementation. Its benefit is that expensive update types are assigned to a smaller subset of samples instead of every rollout group.

\section{Additional Notes on Confidence Diagnostics}

The confidence signal used by the router is computed from token-level entropy and normalized within the batch. We also ran auxiliary GSM8K diagnostics to check whether confidence and entropy behave differently on correct and incorrect responses. These diagnostics are not used for model selection. They serve as sanity checks for the routing signal.

Across the analyzed runs, correct responses often show slightly higher confidence and lower entropy than incorrect responses, though the gap varies by model family and scale. This supports using confidence as a secondary signal rather than as the only routing criterion. Accuracy alone captures whether the current policy can solve the sample; confidence adds information about how stable the model appears when producing those answers.

\section{Router Motivation and Ablations}
\label{app:router_diagnostics}

We conduct a preliminary diagnostic on Qwen3-4B to examine which
optimization recipes are suitable for different rollout states. For each
DAPO-Math-17K prompt, we sample eight rollouts and compute rollout accuracy
$a$ and normalized confidence $c$. We construct behavior-specific 50\%
training subsets and separately apply GRPO and OPSD.

\begin{table}[h]
\centering
\small
\setlength{\tabcolsep}{3pt}
\caption{Diagnostic experiments motivating the behavior-to-recipe mapping.
Results are ID Math / General.}
\label{tab:router_diagnostics}
\begin{tabular}{lccc}
\toprule
Subset & Score & GRPO & OPSD \\
\midrule
Random
& random & 70.9 / 53.8 & 74.8 / 55.6 \\
Recoverable-low
& $(1-a)(1-c)$ & 71.8 / 53.7 & \textbf{76.9 / 56.1} \\
Mixed-correctness
& $a(1-a)$ & \textbf{76.2 / 55.8} & 75.5 / 55.5 \\
Stable-solved
& $ac$ & 70.6 / 52.7 & 72.4 / 53.4 \\
Overconfident-failure
& $(1-a)c$ & 69.4 / 52.5 & 72.7 / 54.0 \\
\bottomrule
\end{tabular}
\end{table}

Mixed-correctness samples favor GRPO, consistent with the availability of
reward contrast within the rollout group, whereas recoverable
low-confidence failures favor OPSD. Stable solved samples show little
benefit from aggressive optimization, while confident failures perform
poorly under both active recipes. These observations motivate the four
routing branches. We therefore view the router as an interpretable
empirical design rather than a theoretically optimal assignment rule.

\section{Additional Router Ablations}
\label{app:router_ablation}

\begin{table}[h]
\centering
\small
\setlength{\tabcolsep}{4pt}
\caption{Ablations of the Self-Routing design on Qwen3-4B.}
\label{tab:router_ablation}
\begin{tabular}{lccc}
\toprule
Variant & ID Math & OOD-V & General \\
\midrule
Self-Routing
& \textbf{80.9} & \textbf{71.1} & \textbf{59.3} \\
Accuracy-only
& 79.6 & 70.4 & 58.1 \\
Confidence-only
& 74.2 & 67.6 & 54.8 \\
w/o conf. calibration
& 79.0 & 70.0 & 57.2 \\
w/o behavior assignment
& 77.8 & 69.0 & 56.4 \\
w/o REG/SKIP
& 79.1 & 70.3 & 56.9 \\
\bottomrule
\end{tabular}
\end{table}

Accuracy provides the strongest routing signal, as accuracy-only routing
remains relatively close to Self-Routing while confidence-only routing
degrades substantially. Confidence nevertheless provides complementary
information, and removing either confidence calibration,
behavior-conditioned assignment, or the conservative REG/SKIP branches
consistently reduces performance.

\section{Additional Baselines and OOD Evaluation}
\label{app:additional_results}

\begin{table}[h]
\centering
\small
\setlength{\tabcolsep}{3pt}
\caption{Additional evaluation on Qwen3-4B. OOD-V is the average over
SATBench, AutoLogi, and LiveCodeBench-v5.}
\label{tab:additional_results}
\begin{tabular}{lccc}
\toprule
Method & ID Math & OOD-V & General \\
\midrule
Base          & 64.4 & 68.2 & 54.3 \\
Naive-GRPO    & 73.0 & 68.5 & 54.2 \\
Naive-OPSD    & 77.3 & 69.5 & 56.6 \\
DAPO-style RL & 75.6 & 69.1 & 55.5 \\
PODS          & 74.8 & 69.1 & 55.4 \\
Self-Routing  & \textbf{80.9} & \textbf{71.1} & \textbf{59.3} \\
\bottomrule
\end{tabular}
\end{table}

\end{document}